\documentclass[conference]{IEEEtran}
\IEEEoverridecommandlockouts

\usepackage{cite}
\usepackage{amsmath,amssymb,amsfonts}
\usepackage{algorithmic}
\usepackage{graphicx}
\usepackage{textcomp}
\usepackage{xcolor}
\def\BibTeX{{\rm B\kern-.05em{\sc i\kern-.025em b}\kern-.08em
    T\kern-.1667em\lower.7ex\hbox{E}\kern-.125emX}}

\usepackage{pifont} 
\usepackage{array}  
\usepackage{booktabs}
\usepackage{multirow}
\usepackage{subcaption}

\begin{document}
\title{An Efficient Self-Supervised Framework for Long-Sequence EEG Modeling}

\author{
\IEEEauthorblockN{Jiazhen Hong}
\IEEEauthorblockA{\textit{Emotiv Research} \\
Melbourne, Australia \\
jiazhen@emotiv.com}
\and
\IEEEauthorblockN{Geoffrey Mackellar}
\IEEEauthorblockA{\textit{Emotiv Research} \\
Sydney, Australia \\
geoff@emotiv.com}
\and
\IEEEauthorblockN{Soheila Ghane}
\IEEEauthorblockA{\textit{Emotiv Research} \\
Melbourne, Australia \\
soheila@emotiv.com}
}

\maketitle
\IEEEpeerreviewmaketitle

\begin{abstract}
Electroencephalogram (EEG) signals generally exhibit low signal-to-noise ratio (SNR) and high inter-subject variability, making generalization across subjects and domains challenging. Recent advances in deep learning, particularly self-supervised learning with Transformer-based architectures, have shown promise in EEG representation learning. However, their quadratic computational complexity increases memory usage and slows inference, making them inefficient for modeling long-range dependencies. Moreover, most existing approaches emphasize either explicit window segmentation of the temporal signal or spectral-only input embedding while neglecting raw temporal dynamics. In this paper, we propose EEGM2, a self-supervised framework that overcomes these limitations. EEGM2 adopts a U-shaped encoder–decoder architecture integrated with Mamba-2 to achieve linear computational complexity, thereby reducing memory usage and improving inference speed. Meanwhile, the selective information propagation mechanism of Mamba-2 enables the model to effectively capture and preserve long-range dependencies in raw EEG signals, where traditional RNN or CNN architectures often struggle. Moreover, EEGM2 employs a self-supervised pre-training objective that reconstructs raw EEG using a combined L1 and spectral (Fourier-based) loss, enhancing generalization by jointly preserving temporal dynamics and spectral characteristics. Experimental results demonstrate that EEGM2 achieves state-of-the-art performance in both short- and long-sequence modeling and classification. Further evaluations show that EEGM2 consistently outperforms existing models, demonstrating strong generalization across subjects and tasks, as well as transferability across domains. Overall, EEGM2 offers an efficient and scalable solution suitable for deployment on resource-constrained brain-computer interface (BCI) devices.
\end{abstract}


\begin{IEEEkeywords}
Brain–computer interface (BCI), Electroencephalography (EEG), Self-supervised learning, State-space models, Mamba-2,  Long-sequence modeling, Representation learning
\end{IEEEkeywords}

\section{Introduction}
Electroencephalography (EEG)-based brain-computer interfaces (BCIs) provide new opportunities to improve the quality of life for individuals with disabilities by enabling control through mental activities \cite{hong2024chatbci}. However, scalp EEG signals are inherently noisy and exhibit high inter-subject variability, posing significant challenges for accurate modeling and generalization~\cite{acharya2013automated}. As a result, traditional EEG analysis methods have primarily focused on subject-specific tasks, such as event-related potential (ERP) detection~\cite{wang2023st} and motor imagery classification~\cite{hong2022deep}, with limited analysis on generalization across subjects or domains.

Deep learning models, particularly self-supervised learning (SSL) approaches, have achieved success in modeling EEG signals across subjects by pre-training on large-scale unlabeled EEG datasets to learn representations, which can then be transferred to downstream tasks \cite{eeg2rep2024}. Transformer-based SSL models for EEG have shown strong performance on short sequence modeling tasks, typically ranging from 2 to 10 seconds \cite{kostas2021bendr, yang2024biot, wangeegpt, eeg2rep2024}. However, as EEG sequence length increases, the computational complexity of Transformer-based models grows quadratically, increasing memory usage during training and inference time \cite{dai2023multichannelsleepnet}, making them inefficient for modeling long sequences of 10,000 or more time steps \cite{gu2021efficiently}.

Moreover, patch-based input embedding methods rely on explicit segmentation using fixed window sizes, which may capture intra-window relationships but often discard inter-window and long-range temporal dependencies~\cite{chen2024eegformer, wangeegpt, jiang2024large, yang2024biot}. Some approaches reduce sequence dimensionality through multiple convolutional downsampling operations~\cite{kostas2021bendr, chien2211maeeg}, further limiting their ability to preserve fine-grained temporal dynamics in long EEG sequences. Other works like BIOT~\cite{yang2024biot} focus exclusively on features in the frequency domain using a fast Fourier transformer, neglecting the temporal structure of the signal. In contrast, models such as EEG2Rep and EEGPT~\cite{eeg2rep2024, wangeegpt} emphasize the alignment of latent representation while overlooking spectral characteristics, thus limiting their ability to generalize across frequency-dependent EEG patterns.



Considering the aforementioned limitations: (1) lack of generalization, (2) high memory consumption, and (3) reliance on explicit temporal segmentation, spectral-only, or temporal-only feature extraction, we  propose \textit{EEGM2},  inspired by two promising architectures: the image reconstruction backbone U-Net~\cite{ronneberger2015unet} and the recently introduced Structured State Space Models (SSMs)~\cite{gu2021efficiently}, particularly Mamba-2~\cite{dao2024transformers}. U-Net, composed of a CNN-based encoder and mirrored decoder, has been widely adopted as the denoising backbone in diffusion models~\cite{croitoru2023diffusion}, owing to its ability to capture both local and global context via skip connections. However, while effective for spatial feature learning, CNN-based architectures often struggle with long-sequence modeling due to limited receptive fields and gradient vanishing issues. In contrast, Mamba~\cite{gu2023mamba} and its successor Mamba-2~\cite{dao2024transformers} leverage selective state propagation to maintain long-range dependencies with linear computational complexity. Mamba-2 further introduces the Structured State-Space Duality (SSD) mechanism, combining a parallel Mamba layer with attention operations. This design achieves a trade-off between modeling fidelity and efficiency, outperforming both the original Mamba and Transformer layers~\cite{dao2024transformers}. Although Mamba-based models have shown strong performance in natural language and vision tasks~\cite{lenz2025jamba, glorioso2024zamba}, applying them to EEG data is non-trivial.

Our proposed EEGM2 is a novel self-supervised framework that integrates a U-shaped encoder–decoder architecture with Mamba-2 blocks. This hybrid design enables EEGM2 to capture both fine-grained and long-range temporal dependencies in raw EEG data, making it a compelling backbone for efficient and scalable EEG representation learning, particularly in long-sequence modeling tasks. The main contributions of EEGM2 are summarized as follows:

\begin{itemize}
    \item \textbf{Low Memory Usage \& Fast Inference Speed:} EEGM2 adopts a hierarchical U-shaped encoder-decoder structure with an integrated Mamba-2-based mediator module, achieving linear computational complexity. This design reduces memory usage and improves inference speed, making it suitable for deployment in resource-constrained environments.

    \item \textbf{Long-sequence EEG modeling:} By incorporating Mamba-2 structured state space models, EEGM2 effectively captures and preserves long-range dependencies in raw EEG signals, where traditional recurrent neural networks (RNNs) and CNNs often struggle.

    \item \textbf{Temporal and spectral feature preservation:} EEGM2 employs a reconstruction-based pretraining objective that combines temporal-domain L1 loss with frequency-domain spectral loss (via Fourier transform). This joint optimization preserves both temporal and spectral characteristics of EEG signals, mitigating the risk of biased representations caused by single-domain reliance.


    \item \textbf{Lightweight alternative variant:} We further introduce \textit{EEGM2 (Light)}, a compact variant that utilizes only the encoder and incorporates a customized representation method for non-linear probing. Despite a reduction in model size by 18$\times$ (from 4.5M to 0.25M parameters), EEGM2 (Light) maintains strong performance, providing an efficient solution for downstream BCI deployments with minimal compromise in accuracy.
\end{itemize}

\section{Methodology}
\begin{figure*}
  \centering
  \includegraphics[width=0.9\linewidth]{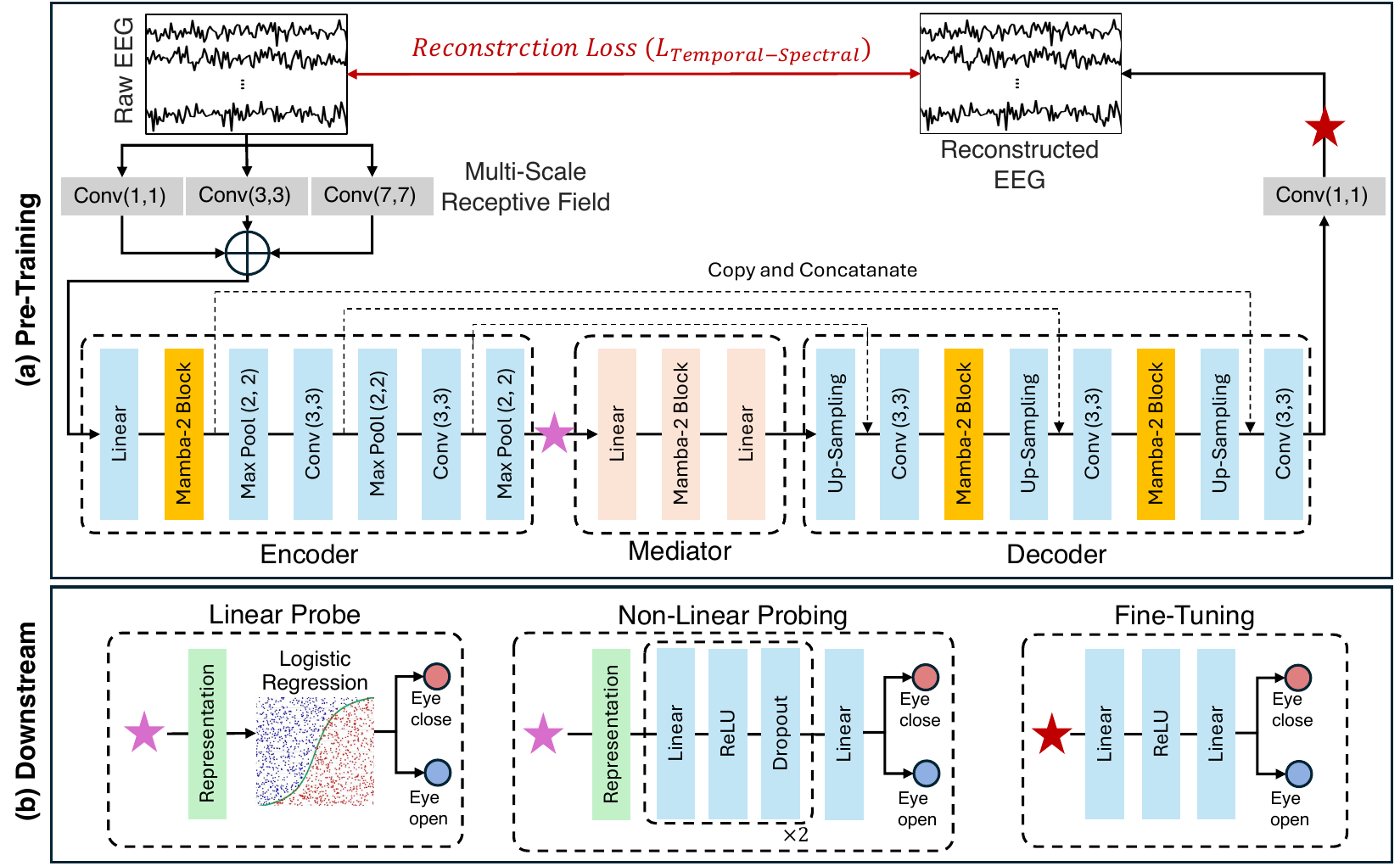}
 \caption{Overview of the EEGM2 framework. (a) Reconstruction-based self-supervised pretraining, where the model learns to reconstruct raw EEG signals using a multi-scale encoder–mediator–decoder architecture supervised by a temporal–spectral loss. No labels are required. (b) Downstream evaluation strategies: linear probing and non-linear probing extract frozen encoder representations (purple star), followed by a logistic regression or MLP classifier, respectively; fine-tuning jointly updates the entire model starting from the red star. The “eye open/close” example represents the binary class labels from the Crowdsourced EEG dataset (Section~\ref{sec:data}).}\label{fig:1}
\end{figure*}

\subsection{Problem Formulation}\label{sec:2.1}
Given a batch of multichannel EEG sequences $\mathbf{X} \in \mathbb{R}^{B \times C_{\text{in}} \times T}$, where $B$ denotes the batch size, $C_{\text{in}}$ is the number of input EEG channels, and $T$ is the number of time points per trial, our goal is to learn robust and generalizable representations of EEG signals through a self-supervised reconstruction task, without relying on any labeled data.

In the pre-training stage (Fig.~\ref{fig:1} (a)), EEGM2 aims to learn a mapping $f_{\theta}: \mathbf{X} \rightarrow \hat{\mathbf{X}}$ that reconstructs the raw EEG input $\mathbf{X}$. To ensure the preservation of both temporal and spectral characteristics, we define the total reconstruction loss as a weighted sum of temporal and spectral terms:
\begin{equation}\label{eq:1}
\mathcal{L}_{\text{reconstruction}} = \alpha \cdot \|\mathbf{X} - \hat{\mathbf{X}}\|_1 + \beta \cdot \|\mathcal{F}(\mathbf{X}) - \mathcal{F}(\hat{\mathbf{X}})\|_2^2,
\end{equation}
where $\mathcal{F}(\cdot)$ denotes the real-valued fast Fourier transform (rFFT), and $\alpha$, $\beta$ are hyperparameters balancing the contribution of temporal and frequency-domain reconstruction errors.

After pre-training, we evaluate the quality of the learned representations on multiple downstream EEG classification tasks. Each dataset provides labeled samples $(\mathbf{X}, y)$, where $\mathbf{X}$ denotes the EEG input and $y$ is the associated class label. This work primarily focuses on binary classification scenarios. We adopt three evaluation strategies (illustrated in Fig.~\ref{fig:1}(b)):
\begin{itemize}
    \item \textit{Linear probing (EEGM2(Linear))}: The pretrained encoder is frozen (purple star), and representations are extracted from a target layer. A logistic regression classifier is trained on these fixed features using the One-vs-Rest strategy.
    
    \item \textit{Non-linear probing (EEGM2(Light))}: Similar to linear probing, the encoder is frozen. However, a lightweight multi-layer perceptron (MLP) with two hidden layers is trained on top of the extracted representations to allow for non-linear decision boundaries.
    
    \item \textit{Fine-tuning (EEGM2(Fine))}: The entire EEGM2 model (starting from the red star) is trained end-to-end alongside a lightweight MLP classification head. All model parameters are updated during training, and convergence is typically achieved within 5 epochs.
\end{itemize}

This evaluation protocol allows EEGM2 to be tested under various practical scenarios and demonstrates its ability to transfer effectively across multiple downstream tasks, including cross-subject, multi-task, and cross-dataset EEG classification tasks.

\subsection{EEGM2 Architecture}
Fig.~\ref{fig:1}(a) illustrates the overall architecture of EEGM2. Given an input EEG sequence $\mathbf{X} \in \mathbb{R}^{B \times C_{\text{in}} \times T}$, the model first applies a multi-scale receptive field module composed of parallel 1D convolutions with kernel sizes 1, 3, and 7. This allows the network to perceive temporal dependencies at multiple scales.

The encoder comprises three stages that progressively increase feature dimensionality while reducing temporal resolution. The first stage applies a linear projection to model inter-channel relationships, followed by a Mamba-2 block that captures fine-grained temporal dynamics. The subsequent stages utilize temporal convolutions and max-pooling operations, enabling deeper layers to efficiently model broader temporal contexts.

The mediator is designed to learn a latent bottleneck representation of the input. It consists of a linear layer, a Mamba-2 block, and a subsequent linear projection. This structure refines inter-channel relationships and encodes temporal dynamics into a compressed latent space.

The decoder mirrors the encoder structure and performs upsampling via parameter-free linear interpolation instead of transposed convolutions, which helps avoid checkerboard artifacts~\cite{sugawara2019checkerboard} and preserves temporal continuity in the reconstructed EEG. Each upsampled feature map is further processed by a Mamba-2 block to restore long-range dependencies. Skip connections from the encoder are concatenated with the corresponding decoder layers to retain high-resolution features. Finally, a $1 \times 1$ convolutional layer is applied to generate the output reconstruction.

\subsection{Mamba-2 Block}
Each Mamba-2 block in EEGM2 is equipped with LayerNorm and residual connections to enhance training stability and performance. Mamba-2~\cite{dao2024transformers} is a structured state-space model (SSM) designed for efficient long-sequence modeling. Compared to Transformer-based models, Mamba-2 achieves linear time complexity while preserving the ability to capture long-range dependencies, making it well-suited for EEG data. Its structured state-space duality (SSD) enables both parallel computation and selective information propagation, improving computational efficiency and modeling scalability.

Since Mamba-2 requires input sequences of shape $(T, C)$, each intermediate representation in EEGM2 is transposed from $(B, C, T)$ to $(B, T, C)$ before being passed into a Mamba-2 block, and then reverted back to $(B, C, T)$ afterward. This transposition ensures compatibility while preserving spatial-temporal alignment across channels and time steps. Formally, given an input sequence $\mathbf{x} \in \mathbb{R}^{T \times d}$, Mamba-2 evolves a latent state via:
\begin{equation}
    \mathbf{h}_t = A_t \mathbf{h}_{t-1} + B_t \mathbf{x}_t,
\end{equation}
\begin{equation}
    \mathbf{y}_t = C_t \mathbf{h}_t,
\end{equation}
where $t \in \{1, \dots, T\}$ is the time step, and $A_t$, $B_t$, and $C_t$ are parameterized matrices that are dynamically updated during training. In contrast to traditional recurrent models, Mamba-2 employs a selective update mechanism that improves expressiveness and robustness in modeling long-range dependencies in EEG signals.

\subsection{Customized Representation Extraction}\label{sec:2.4}
To enable efficient inference for downstream classification, especially with long EEG sequences, we design a customized representation extraction approach, highlighted in green in Fig.~\ref{fig:1}(b). Features are captured from the encoder using a forward hook, avoiding direct flattening or global pooling.

Given an input EEG sequence $\mathbf{X} \in \mathbb{R}^{B \times C \times T}$, we register a forward hook on a target encoder module to obtain hidden representations $\mathbf{F} \in \mathbb{R}^{B \times C \times T'}$ during inference. We then summarize $\mathbf{F}$ along the temporal axis using descriptive statistics: minimum, maximum, mean, standard deviation, and quantiles (0.05–0.95). The resulting vector is defined as:
\begin{equation}
    \mathbf{z} = [\min, \max, \mu, \sigma, Q_{0.05}, Q_{0.25}, Q_{0.5}, Q_{0.75}, Q_{0.95}],
\end{equation}
yielding a compact feature tensor $\mathbf{z} \in \mathbb{R}^{B \times C \times 9}$, where each of the nine dimensions encodes a temporal statistic per channel. For non-linear probing, this tensor is directly fed into the MLP. For linear classification, it is flattened into a vector of shape $\mathbb{R}^{B \times (9 \cdot C)}$. A visualization of the learned representations is provided in Section~\ref{sec:tSNE}.


\subsection{Temporal-Spectral Loss Function}
To preserve both temporal dynamics and spectral characteristics of EEG signals during pre-training, we adopt a dual-domain reconstruction loss (Equation~\ref{eq:1}). The temporal loss is defined as the Mean Absolute Error (L1 loss) between predicted and target waveforms. Compared to L2 (Mean Squared Error), L1 avoids excessive penalization of large deviations, resulting in better generalization~\cite{mazilu2011l1}. Such deviations are often caused by noise and artifacts in EEG signals, making L1 a more robust choice for EEG modeling. However, due to its non-smooth gradients at zero, L1 loss may slow convergence. To mitigate this, we apply a OneCycle learning rate schedule (Section~\ref{sec:setup}) to stabilize and accelerate training.

The spectral loss imposes an auxiliary constraint in the frequency domain by enforcing similarity between the predicted and ground truth amplitude spectra. We apply a real-valued Fast Fourier Transform (rFFT) along the temporal axis and minimize the mean squared magnitude difference between the predicted and ground truth spectra across all frequencies. Unlike approaches that focus on predefined frequency bands, our formulation reconstructs the entire spectral profile. Although it does not explicitly isolate alpha, beta, or theta bands, it encourages the retention of all frequency components present in the input, thus complementing the time-domain loss and promoting the reconstruction of waveform integrity and oscillatory features critical for downstream decoding.

\section{Experiments}
\subsection{Dataset}\label{sec:data}
\begin{table}
\centering
\caption{Overview of the datasets used in this study.}
\label{tab:1}
\begin{tabular}{l|c|c|c|c}
\hline
\textbf{Datasets}  & \textbf{Chan.} &  \textbf{Sub.} & \textbf{Samples}  & \textbf{Seq. Length} \\ \hline
\textbf{TUAB} - 10 s          & 16 &  2383     & 409,455                           & 1280              \\ 
\textbf{TUAB} - 30 s          & 16 &  2383     & 135,702                            & 3840              \\ 
\textbf{TUAB} - 60 s          & 16 &  2383     & 56,290                              & 7680              \\ 
\textbf{TUAB} - 100 s         & 16 &  2383     & 39,810                              & 12800              \\ \hline
\textbf{Crowdsourced}      & 14 &  13       & 12,296                             & 256               \\ 
\textbf{STEW}           & 14 &  48       & 28,512                             & 256               \\ 
\textbf{DriverDistraction}   & 14 &  17       & 66,197                             & 256               \\ 
\textbf{Alpha}                & 14 & 59 & 11,866                             & 256               \\ 
\textbf{Attention}            & 14 & 27 & 21,894                             & 256               \\ \hline
\end{tabular}
\end{table}

Table~\ref{tab:1} summarizes the datasets used in this study, including the TUH Abnormal EEG Corpus (TUAB)~\cite{lopez2015automated} and five Emotiv datasets covering diverse real-world scenarios.

\noindent\textbf{TUAB:} The TUAB dataset is a large-scale clinical EEG corpus in which each recording is labeled as either normal or abnormal based on neurologist reports. It is widely used for automated abnormal EEG detection. TUAB provides 16-channel EEG signals sampled at 200 Hz, which we downsample to 128 Hz to match the Emotiv datasets. Following BIOT~\cite{yang2024biot}, we adopt the same training, validation, and test set split strategy to ensure fair comparability with prior work. Benefiting from its long-term monitoring (LTM) recordings, TUAB contains segments spanning several hours. Previous Transformer-based SSL methods typically divide TUAB into 2–10 second windows due to computational constraints~\cite{eeg2rep2024, wangeegpt, yang2024biot}. To evaluate EEGM2’s ability to model long EEG sequences, we experiment with four sequence lengths: 10, 30, 60, and 100 seconds.

\noindent\textbf{Emotiv:} The Emotiv datasets were collected using the Emotiv Epoc headset, which records 14-channel EEG signals at 128 Hz. All signals were bandpass filtered and segmented into 2 second windows (256 time steps). We evaluate EEGM2 on five Emotiv datasets drawn from real-world applications: (1) Crowdsourced Eye Open/Close Detection~\cite{williams2023crowdsourced}, (2) STEW, a workload estimation dataset~\cite{lim2018stew}, (3) Driver Distraction Detection, (4) Eye Open/Close Detection based on Alpha Band Activity, and (5) Attention State Classification. Among these, the first two are publicly available, while the remaining three are proprietary datasets provided by Emotiv. To evaluate cross-subject generalization, all Emotiv datasets are split in a \emph{subject-wise} manner: EEG recordings from each subject are assigned exclusively to either the training, validation, or test set. In other words, no EEG data from the same individual appears in more than one split. This setup introduces significant inter-subject variability and allows us to assess EEGM2’s generalization to entirely unseen subjects. Additionally, the datasets span multiple Emotiv headset types, each with distinct signal-noise characteristics. This diversity further challenges the model’s adaptability to different hardware conditions. More details on dataset statistics and preprocessing are provided in Appendix~\ref{app:1}.

\subsection{Implementation Details}
\subsubsection{\textbf{Pre-Training \& Downstream Tasks}} \label{sec:setup}
The experiment consists of two stages. In the first stage, the model is pretrained in a self-supervised manner for 500 epochs with a batch size of 64, using only unlabeled data and optimizing the reconstruction loss. We employ the AdamW optimizer with an initial learning rate of $2.5 \times 10^{-4}$ and a weight decay of $1 \times 10^{-2}$. A OneCycle learning rate schedule~\cite{smith2019super} is applied, with a maximum learning rate of $5 \times 10^{-4}$ and a minimum of approximately $3.13 \times 10^{-5}$, following a cosine annealing strategy. The learning rate is warmed up during the first 30\% of training steps, starting at $\frac{\text{max\_lr}}{10}$ and gradually decaying to $\frac{\text{max\_lr}}{10000}$ by the end of training.

In the second stage, we evaluate the pretrained encoder on supervised downstream classification tasks using three strategies (see Fig.~\ref{fig:1}(b) and Section~\ref{sec:2.1}). For all downstream tasks, we adopt the cross-entropy loss and the AdamW optimizer. Each classification model converges within 5 epochs. To ensure robustness, every experiment is repeated three times, and we report the mean and standard deviation across all results presented in this paper. In addition to classification accuracy, we report \textit{Balanced Accuracy (ACC)}, which computes the average recall across all classes and serves as a more reliable metric for imbalanced datasets. We also report the \textit{Area Under the Receiver Operating Characteristic Curve (AUROC)}, which quantifies the model’s ability to distinguish between classes by summarizing the ROC curve into a single scalar value.

Both pre-training and downstream training are performed using 32-bit mixed precision on a single NVIDIA RTX 6000 Ada GPU. 

\subsubsection{EEGM2 Variant Settings in Ablation Study} \label{sec:variant}
To assess the contribution of each component, we design five ablation variants of EEGM2, each modifying or removing a specific module. The six configurations evaluated are as follows:

\begin{itemize}
    \item \textbf{EEGM2}: The full model incorporating all components shown in Fig.~\ref{fig:1}, serving as the reference configuration.
    \item \textbf{EEGM2-S1}: A variant without the multi-scale receptive field input embedding.
    \item \textbf{EEGM2-S2}: A variant that replaces the temporal-spectral loss with L1 loss only.
    \item \textbf{EEGM2-S3}: A variant in which all Mamba-2 blocks are replaced with Mamba-1 blocks.
    \item \textbf{EEGM2-S4}: Similar to EEGM2-S3, but additionally removes the multi-scale receptive field.
    \item \textbf{EEGM2-S5}: A variant in which the Mamba-2 blocks are replaced with Transformer blocks.
\end{itemize}

These variants allow us to systematically isolate the effects of the input embedding, loss function, Mamba blocks, and backbone architecture on overall model performance.

\section{Results}
In this section, we first demonstrate the representation capability of EEGM2 through visualization in Section~\ref{sec:tSNE}. We then evaluate EEGM2's effectiveness in modeling long EEG sequences and present ablation studies on both pre-training and downstream tasks in Section~\ref{sec:4.1}. Section~\ref{sec:4.2} examines the generalization of EEGM2 through in-domain downstream classification and investigates its transferability via cross-domain downstream classification. Finally, Section~\ref{sec:4.3} compares EEGM2 with recent Transformer-based models in terms of memory usage and inference speed under varying sequence lengths using simulation experiments.

\subsection{t-SNE Visualization of EEGM2 Representation}\label{sec:tSNE}
\begin{figure}
  \centering
  \begin{subfigure}[b]{0.49\linewidth} 
    \centering
    \includegraphics[width=\linewidth]{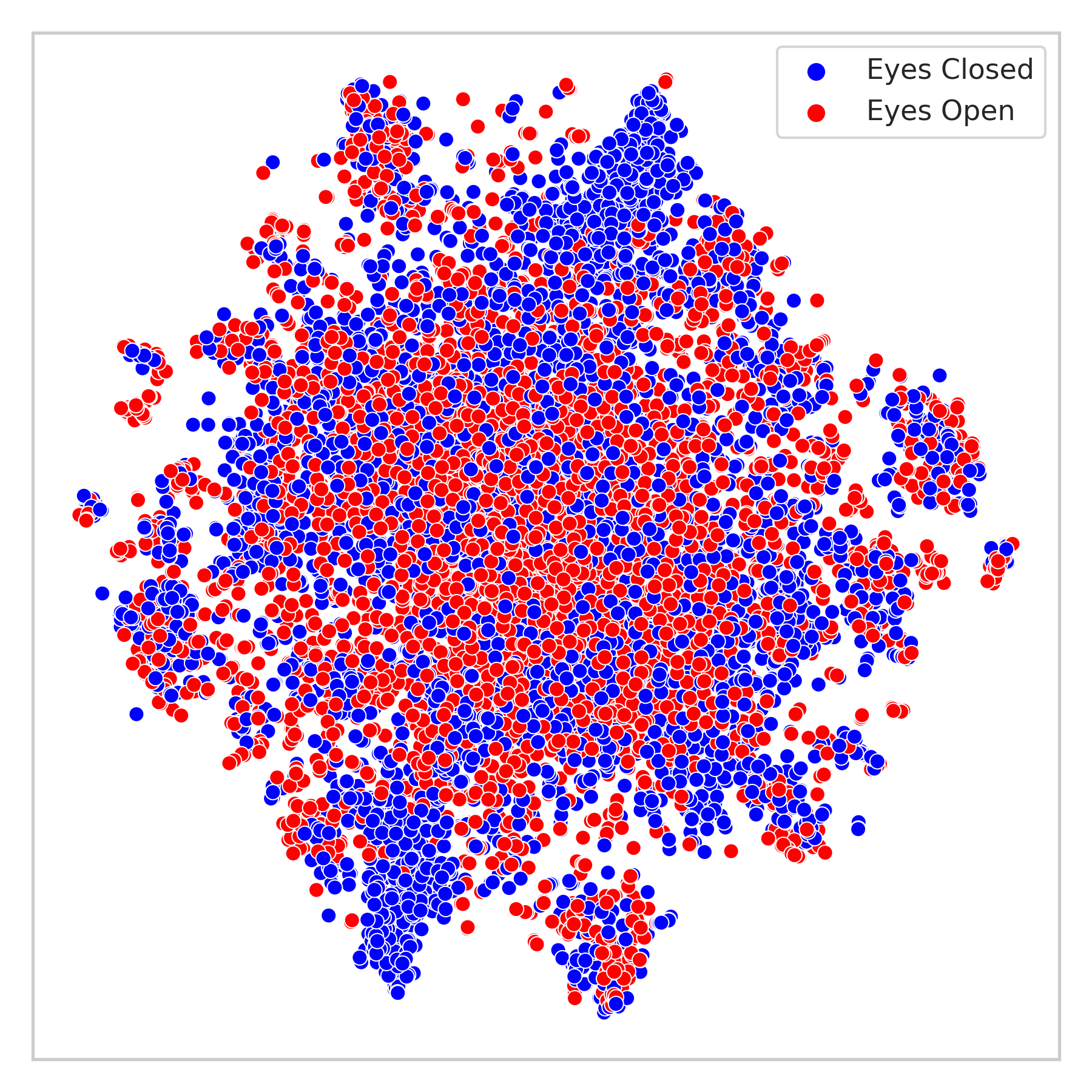}
    \caption{Raw Signal}
    \label{fig:2a}
  \end{subfigure}
  \hfill 
  \begin{subfigure}[b]{0.49\linewidth}
    \centering
    \includegraphics[width=\linewidth]{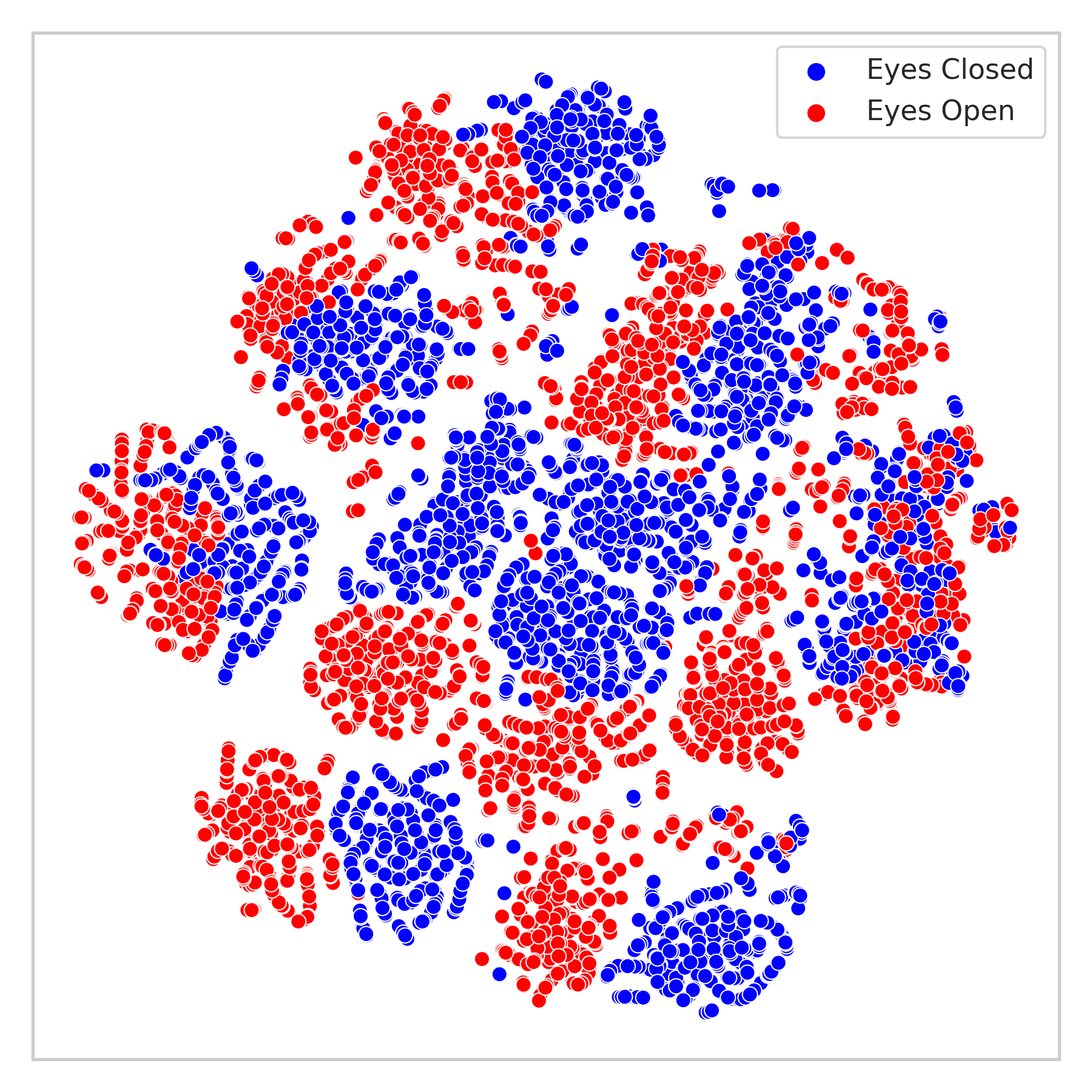}
    \caption{EEGM2 Representation}
    \label{fig:2b}
  \end{subfigure}
\caption{Comparison of 2D t-SNE projections of (a) raw mean EEG features and (b) EEGM2-learned representations on the Crowdsourced EEG dataset.}
  \label{fig:2}
\end{figure}

Fig.~\ref{fig:2} presents a two-dimensional visualization using t-distributed Stochastic Neighbor Embedding (t-SNE)~\cite{van2008visualizing}, comparing raw EEG signal features (Fig.~\ref{fig:2}a) and EEGM2 representations after pre-training (Fig.~\ref{fig:2}b). The representations are extracted from the encoder layer using the proposed customized representation extraction method described in Section~\ref{sec:2.4}. The evaluation is conducted on the test set of the Crowdsourced EEG dataset. t-SNE projects high-dimensional EEG features into two-dimensional space for visual inspection. As shown in Fig.~\ref{fig:2}, EEGM2 yields more compact and well-separated class clusters compared to raw features, demonstrating its ability to extract class-discriminative and semantically meaningful representations and confirming its effectiveness in learning robust EEG representations.

\subsection{Long-Sequence EEG Modeling} \label{sec:4.1}
\begin{table*} 
\centering
\caption{Performance comparison of EEGM2 variants in pre-training for long-sequence EEG modeling.}
\label{tab:2}
\begin{tabular}{|l|c|c|c|c|c|}
\hline
\textbf{Models}                            & \textbf{Mamba-2 Block} & \textbf{Temporal-Spectral Loss} & \textbf{Multi-Scale} & \textbf{ACMSE} & \textbf{Averaged Training Time} \\ \hline
EEGM2                              & \ding{51}              & \ding{51}                    & \ding{51}                             & 6.87e-13       & 299.43 seconds/epoch                \\ \hline
EEGM2-S1       & \ding{51}              & \ding{51}                    & \ding{55}                             & 9.24e-12       & 296.45 seconds/epoch                   \\ \hline
EEGM2-S2        & \ding{51}              & \ding{55} (L1 Loss)          & \ding{51}                             & 7.74e-12       & 300.12 seconds/epoch                   \\ \hline
EEGM2-S3   & Mamba-1                & \ding{51}                    & \ding{51}                             & 6.95e-13       & 435.95 seconds/epoch                   \\ \hline
EEGM2-S4                       & Mamba-1                & \ding{51}                    & \ding{55}                             & 1.06e-11       & 438.05  seconds/epoch                  \\ \hline
EEGM2-S5  & Transformer              & \ding{51}                    & \ding{51}                             & Out of Memory  & Out of Memory          \\ \hline
\end{tabular}
\end{table*}

Here, we investigate the long-sequence modeling capability of EEGM2 and conduct an ablation study using five variant configurations to evaluate the contribution of each architectural component. Table~\ref{tab:2} summarizes the self-supervised reconstruction performance of EEGM2 and its variants on long EEG sequences from the TUAB dataset, with each input spanning 100 seconds (12{,}800 time steps). All experiments are conducted under identical settings on a single NVIDIA RTX 6000 Ada GPU. To assess modeling quality and efficiency, we report two metrics: (1) \textit{Averaged Channel-wise Mean Squared Error (ACMSE)}, computed by first calculating the MSE for each EEG channel and then averaging across all channels, and (2) average training speed measured in epochs per second. These metrics provide insights into both reconstruction accuracy and computational efficiency, allowing us to quantify the impact of each component in EEGM2.

As shown in Table~\ref{tab:2}, EEGM2 achieves the lowest ACMSE, demonstrating strong capability in modeling long-sequence EEG data. Notably, EEGM2-S5, where the Mamba-2 blocks are replaced by Transformer blocks, fails to process 100-second sequences due to the Transformer's quadratic computational complexity, resulting in out-of-memory (OOM) errors. This underscores the advantage of Mamba-2’s linear scaling, which enables EEGM2 to handle long sequences efficiently while maintaining superior reconstruction fidelity. Additional analysis of memory usage is provided in Section~\ref{sec:4.3}. When comparing EEGM2 to EEGM2-S4, we observe that although the improvement in ACMSE is modest, the Mamba-2 blocks lead to a notable reduction in training time compared to the Mamba-1 variant, demonstrating enhanced computational efficiency. Comparisons between EEGM2 and EEGM2-S1, as well as between EEGM2-S4 and EEGM2-S5, show that the multi-scale receptive field introduces only a slight increase in training time while significantly improving reconstruction accuracy, demonstrating its effectiveness. Moreover, EEGM2-S2, which omits the temporal-spectral loss, exhibits comparable training speed but suffers from unstable training behavior. Additional visualization and insights into training time are provided in Appendix~\ref{app:2}.

\begin{table}
\centering
\caption{Performance Comparison of EEGM2 Variants in Long-Sequence EEG Downstream Task.}
\label{tab:3}
\begin{tabular}{lccc}
\toprule
\textbf{Models} & \textbf{Balanced ACC} & \textbf{AUROC} \\
\midrule
\textbf{EEGM2}    & \textbf{81.08} & \textbf{0.8869}   \\ \hline
\multicolumn{3}{l}{\textbf{w/o multi-scale}} \\ \hline
EEGM2-S1 & 79.06 ($\downarrow$ 2.02) & 0.8546 ($\downarrow$ 0.03) \\
\hline
\multicolumn{3}{l}{\textbf{w/o temporal-spectral loss}} \\ \hline
EEGM2-S2 & 77.01 ($\downarrow$ 4.07) & 0.8247 ($\downarrow$ 0.06)   \\
\hline
\multicolumn{3}{l}{\textbf{w/o mamba-2 block}} \\ \hline
EEGM2-S3 & 76.08 ($\downarrow$ 5.00) & 0.8176 ($\downarrow$ 0.07)  \\
EEGM2-S4 & 75.38 ($\downarrow$ 5.70) & 0.8339 ($\downarrow$ 0.05)  \\ 
EEGM2-S5 & out of memory & out of memory \\ \hline
\end{tabular}
\end{table}

Table~\ref{tab:3} presents an ablation study evaluating the performance of EEGM2 and its variants on the long-sequence EEG downstream task, using representations learned from the corresponding pretrained models analyzed in Table~\ref{tab:2}. Removing the multi-scale receptive field input embedding results in the smallest performance drop, with a 2.02\% decrease in balanced accuracy, indicating its beneficial role in capturing raw EEG features across multiple temporal scales. Excluding the temporal-spectral loss leads to a larger 4.07\% decline, underscoring its importance in preserving spectral information and mitigating noise. Replacing the Mamba-2 block with Mamba-1 causes a notable 5.00\% reduction, highlighting the superior capability of Mamba-2 in modeling long-range temporal dependencies. The greatest performance degradation is observed in EEGM2-S4, where both the multi-scale input embedding is removed and the Mamba-2 block is replaced with Mamba-1, resulting in a 5.70\% decrease in balanced accuracy. These results demonstrate the complementary and critical contributions of these components to EEGM2’s overall performance.

\subsection{Generalization \& Transferability Analysis}
\label{sec:4.2}
In this section, we evaluate the generalization ability of EEGM2 through a series of in-domain and cross-domain experiments. For in-domain analysis, we assess EEGM2 using various sequence lengths on the TUAB dataset and across multiple tasks on the Emotiv datasets, employing a subject-wise split strategy to evaluate cross-subject generalization. For cross-domain analysis, we examine the transferability of EEGM2 by pre-training the model on one dataset and evaluating downstream tasks on a different dataset, thereby demonstrating the robustness and adaptability of the learned representations.

\subsubsection{\textbf{In-domain (TUAB)}}
\begin{table}
\centering
\caption{Performance comparison between EEGM2 and state-of-the-art models on the TUAB dataset across varying sequence lengths.}
\label{tab:4}
\resizebox{\linewidth}{!}{
\begin{tabular}{l|c|c}
\toprule
\textbf{Models} & \textbf{Balanced ACC} & \textbf{AUROC} \\ \hline

\multicolumn{3}{l}{\textbf{10 Seconds:}} \\ \hline
CNN-LSTM \cite{li2022motor, wangeegpt}       & 78.48$\pm$0.38          & 0.8569$\pm$0.0051 \\ 
CNNTransformer \cite{peh2022transformer, wangeegpt}              & 77.77$\pm$0.22          & 0.8461$\pm$0.0013 \\ 
BIOT \cite{yang2024biot}             & 79.59$\pm$0.57          & 0.8815$\pm$0.0043 \\ 
EEGPT \cite{wangeegpt}                       & 79.83$\pm$0.30          & 0.8718$\pm$0.0050 \\
MAEEG \cite{chien2211maeeg, eeg2rep2024}                  &   77.56$\pm$3.56      & 0.8656$\pm$0.0333 \\ 
BENDR \cite{kostas2021bendr, eeg2rep2024}                     &  76.96$\pm$3.98     & 0.8397$\pm$0.0344 \\ 
EEG2Rep \cite{eeg2rep2024}             &   80.52$\pm$2.22     & 0.8843$\pm$0.0309 \\ \hline
EEGM2 (Light)  & 79.14$\pm$0.21 & 0.8559$\pm$0.00  \\
EEGM2 (Fine) & \textbf{80.87}$\pm$0.54 & \textbf{0.8864}$\pm$0.00\\ \hline

\multicolumn{3}{l}{\textbf{30 Seconds:}} \\ \hline
EEGM2 (Light)  & 78.97$\pm$0.25 & 0.8575$\pm$0.00  \\
EEGM2 (Fine) & \textbf{81.71}$\pm$0.12 & \textbf{0.8932}$\pm$0.00 \\ \hline

\multicolumn{3}{l}{\textbf{60 Seconds:}} \\ \hline
EEGM2 (Light)  & 76.94$\pm$0.33 & 0.8257$\pm$0.00  \\
EEGM2 (Fine)  & \textbf{80.68}$\pm$0.45 & \textbf{0.8803}$\pm$0.00 \\ \hline

\multicolumn{3}{l}{\textbf{100 Seconds:}} \\ \hline
EEGM2 (Light)  & 74.57$\pm$0.27 & 0.7986$\pm$0.00    \\ 
EEGM2 (Fine) & \textbf{81.08}$\pm$0.28 & \textbf{0.8869}$\pm$0.00 \\ \hline
\end{tabular}
}
\end{table}

Most prior works on the TUAB dataset focus on short-sequence settings (e.g., 10 seconds) due to memory constraints. Studies such as BIOT~\cite{yang2024biot}, EEGPT~\cite{wangeegpt}, and EEG2Rep~\cite{eeg2rep2024} have demonstrated the effectiveness of self-supervised learning for EEG representation learning, achieving high AUROC scores using CNN-LSTM or Transformer-based architectures. However, these methods often struggle with modeling long-range dependencies due to the quadratic complexity of self-attention or gradient vanishing issues. Benefiting from the integration of the Mamba-2 block, EEGM2 overcomes these limitations. To assess its capacity for modeling long EEG sequences, we conduct experiments on the TUAB dataset using durations ranging from 10 to 100 seconds, corresponding to sequence lengths from 1,280 to 12,800 samples. 

Table~\ref{tab:4} reports the balanced accuracy and AUROC of EEGM2 and state-of-the-art models across four durations: 10, 30, 60, and 100 seconds. For each duration, the best performance is highlighted in bold. We evaluate two configurations of EEGM2: EEGM2 (Light), a lightweight variant with only 0.25M parameters, and EEGM2 (Fine), the full model with 4.5M parameters. For fair comparison, we include the best results reported by each baseline. Note that BENDR, MAEEG, and EEG2Rep report accuracy instead of balanced accuracy, which may overestimate performance on this imbalanced dataset. EEGM2 (Fine) consistently achieves state-of-the-art results across all durations. At 10 seconds, it achieves a balanced accuracy of 80.87\% and an AUROC of 0.8864, outperforming all baselines. Remarkably, its performance improves as the sequence length increases, peaking at 30 seconds with a balanced accuracy of 81.71\% and AUROC of 0.8932. Even at 100 seconds, it maintains strong performance (81.08\%, 0.8869), highlighting its robustness in modeling long-range temporal dependencies. In contrast, EEGM2 (Light) delivers competitive performance at shorter durations (e.g., 79.14\% balanced accuracy and 0.8559 AUROC at 10 seconds), but its performance degrades as the sequence length increases. This decline is likely due to two factors: (1) the customized representation extraction method (Section~\ref{sec:2.4}), which may lose information over long sequences; and (2) the absence of the mediator and decoder modules, which limits its capacity to model high-level temporal features. In comparison, EEGM2 (Fine), with three Mamba-2 blocks across the mediator and decoder, better captures long-range dependencies and delivers consistently superior downstream performance.

\subsubsection{\textbf{In-domain (Emotiv)}}
\begin{table*}
\centering
\caption{Performance comparison of EEGM2 and state-of-the-art models on downstream classification tasks using the Emotiv datasets.}\label{tab:5}
\resizebox{\textwidth}{!}{
\begin{tabular}{l cc cc cc cc cc}
\toprule
\multirow{2}{*}{\textbf{Models}} & \multicolumn{2}{c}{\textbf{Crowdsourced}} & \multicolumn{2}{c}{\textbf{DriverDistraction}} & \multicolumn{2}{c}{\textbf{STEW}} & \multicolumn{2}{c}{\textbf{Alpha}} & \multicolumn{2}{c}{\textbf{Attention}} \\ 
\cline{2-11} 
& \multicolumn{1}{c}{\textbf{ACC}} & \textbf{AUROC} & \multicolumn{1}{c}{\textbf{ACC}} & \textbf{AUROC} & \multicolumn{1}{c}{\textbf{ACC}} & \textbf{AUROC} & \multicolumn{1}{c}{\textbf{ACC}} & \textbf{AUROC} & \multicolumn{1}{c}{\textbf{ACC}} & \textbf{AUROC} \\ 
\hline
MAEEG \cite{chien2211maeeg} & \multicolumn{1}{c}{86.75$\pm$3.50} & 0.8621$\pm$0.03 & \multicolumn{1}{c}{74.58$\pm$2.16} & 0.6079$\pm$0.03 & \multicolumn{1}{c}{72.46$\pm$3.67} & 0.7250$\pm$0.03 & \multicolumn{1}{c}{69.18$\pm$1.54} & 0.7949$\pm$0.02 & \multicolumn{1}{c}{82.61$\pm$0.01} & 0.5282$\pm$0.03 \\ 

BENDR \cite{kostas2021bendr} & \multicolumn{1}{c}{83.78$\pm$2.35} & 0.8380$\pm$0.03 & \multicolumn{1}{c}{74.31$\pm$2.38} & 0.5986$\pm$0.03 & \multicolumn{1}{c}{69.74$\pm$2.11} & 0.6977$\pm$0.02 & \multicolumn{1}{c}{65.75$\pm$2.50} & 0.6764$\pm$0.02 & \multicolumn{1}{c}{76.93$\pm$4.04} & 0.5464$\pm$0.04 \\ 

EEG2Rep \cite{eeg2rep2024} & \multicolumn{1}{c}{94.13$\pm$2.11} & 0.9413$\pm$0.02 & \multicolumn{1}{c}{80.07$\pm$2.63} & 0.6614$\pm$0.02 & \multicolumn{1}{c}{73.60$\pm$1.47} & 0.7440$\pm$0.02 & \multicolumn{1}{c}{73.10$\pm$2.76} & 0.8118$\pm$0.07 & \multicolumn{1}{c}{75.41$\pm$3.20} & 0.6635$\pm$0.03 \\ 

BIOT \cite{yang2024biot} & \multicolumn{1}{c}{87.95$\pm$3.52} & 0.8778$\pm$0.03 & \multicolumn{1}{c}{74.34$\pm$3.57} & 0.6121$\pm$0.04 & \multicolumn{1}{c}{69.88$\pm$2.15} & 0.7011$\pm$0.03 & \multicolumn{1}{c}{70.72$\pm$1.32} & 0.7698$\pm$0.03 & \multicolumn{1}{c}{73.59$\pm$7.24} & 0.6326$\pm$0.13 \\ 

\hline
EEGM2(Linear) & \multicolumn{1}{c}{84.84$\pm$0.05} & 0.9185$\pm$0.00 & \multicolumn{1}{c}{73.99$\pm$0.06} & 0.7046$\pm$0.00 & \multicolumn{1}{c}{68.59$\pm$0.00} & 0.7341$\pm$0.00 & \multicolumn{1}{c}{66.75$\pm$0.13} & 0.7971$\pm$0.00 & \multicolumn{1}{c}{76.05$\pm$0.02} & 0.6479$\pm$0.00 \\ 

EEGM2(Light)                                  & \multicolumn{1}{c}{86.13$\pm$0.21}          & 0.9245$\pm$0.01          & \multicolumn{1}{c}{\textbf{81.11$\pm$0.13}} & 0.6825$\pm$0.01          & \multicolumn{1}{c}{70.24$\pm$0.69}          & 0.7523$\pm$0.01                      & \multicolumn{1}{c}{{75.69$\pm$1.20}}    & { 0.8563$\pm$0.02}                & \multicolumn{1}{c}{\textbf{82.81$\pm$0.35}} & { 0.6708$\pm$0.01}    \\ 
EEGM2(Scratch)                                   & \multicolumn{1}{c}{84.19$\pm$3.83}          & 0.9302$\pm$0.02          & \multicolumn{1}{c}{73.44$\pm$2.93}          & 0.6445$\pm$0.01          & \multicolumn{1}{c}{72.39$\pm$2.39}          & {0.7891$\pm$0.00}                & \multicolumn{1}{c}{68.12$\pm$3.70}          & 0.8137$\pm$0.06                      & \multicolumn{1}{c}{76.58$\pm$1.05}          & 0.6262$\pm$0.06         \\ 

EEGM2(Fine) & \multicolumn{1}{c}{\textbf{94.51$\pm$1.31}} & \textbf{0.9881$\pm$0.00} & \multicolumn{1}{c}{76.54$\pm$1.21} & \textbf{0.7097$\pm$0.01} & \multicolumn{1}{c}{\textbf{74.26$\pm$1.48}} & \textbf{0.7901$\pm$0.02} & \multicolumn{1}{c}{\textbf{77.49$\pm$4.27}} & \textbf{0.8856$\pm$0.02} & \multicolumn{1}{c}{79.14$\pm$3.15} & \textbf{0.6885$\pm$0.01} \\ 
\hline
\bottomrule
\end{tabular}
}
\end{table*}

Table~\ref{tab:5} presents the performance of EEGM2 and state-of-the-art baselines on downstream tasks using the Emotiv datasets, evaluating both model stability and cross-subject generalization. Classification accuracy (ACC) and AUROC are reported, with the best results for each dataset highlighted in bold. To better assess the representation learning capability of EEGM2, in addition to EEGM2(Linear), EEGM2(Light), and EEGM2(Fine), we include EEGM2(Scratch) as a baseline trained from random initialization without pretaining. Additionally, we reproduce several recent Transformer-based models, including MAEEG~\cite{chien2211maeeg}, BENDR~\cite{kostas2021bendr}, EEG2Rep~\cite{eeg2rep2024}, and BIOT~\cite{yang2024biot}, under identical experimental settings for fair comparison.

As shown in Table~\ref{tab:5}, EEGM2(Fine) and EEGM2(Light) consistently achieve top performance across all datasets, surpassing both EEGM2(Scratch) and the state-of-the-art models. Notably, EEGM2(Light), despite using 18$\times$ fewer parameters than both EEGM2(Scratch) and (Fine), EEGM2(Light) still outperforms them on the DriverDistraction and Attention datasets in accuracy. This demonstrates the effectiveness of the pretrained encoder in learning robust representations and highlights EEGM2(Light) as an efficient alternative for short-duration (e.g., 2-second) EEG classification. Class imbalance remains a significant challenge in EEG classification. For instance, in the severely imbalanced DriverDistraction dataset, EEGM2(Fine) achieves the highest AUROC (0.7097), despite a lower accuracy of 76.54\%. This underscores the importance of AUROC as a more reliable evaluation metric under imbalance and further highlights EEGM2’s ability to extract robust, class-discriminative EEG features rather than relying on majority-class predictions. Overall, these results confirm EEGM2’s strong generalization and transfer capabilities across diverse subject-wise tasks, outperforming the state-of-the-art models.

\subsubsection{\textbf{Cross-Domain}}
\begin{table*}
\centering
\caption{EEGM2 performance on in-domain and cross-domain transfer settings, pre-trained on the DriverDistraction dataset.}
\label{tab:6}
\begin{tabular}{l cc cc cc cc cc} 
\toprule
\multirow{2}{*}{\textbf{Initilization}} & \multicolumn{2}{c}{\textbf{Crowdsourced}}            & \multicolumn{2}{c}{\textbf{STEW}}                          & \multicolumn{2}{c}{\textbf{Alpha}}                         & \multicolumn{2}{c}{\textbf{Attention}} & \multicolumn{2}{c}{\textbf{Average}}          \\ \cline{2-11}
  & \textbf{Acc}   & \textbf{AUROC} & \textbf{Acc}   & \textbf{AUROC}        & \textbf{Acc}   & \textbf{AUROC}        & \textbf{ACC}   & \textbf{AUROC}  & \textbf{ACC}   & \textbf{AUROC}  \\ \hline
Random & 84.19 & 0.9302 & 72.96 & 0.7842&	68.12 &	0.8137&	76.58&	0.6262&	75.46	& 0.7886  \\ 
Cross-domain &      ($\uparrow$ 8.33) &	($\uparrow$ 0.0506) &	($\uparrow$ 0.98) &	($\uparrow$ 0.0129) &	($\uparrow$ 1.61) &	($\uparrow$ 0.0597) &	($\uparrow$ 0.07) &	($\downarrow$ 0.0153) &	($\uparrow$ 2.75) &	($\uparrow$ 0.0270)     \\

In-domain   & ($\uparrow$ 10.32)  &	($\uparrow$ 0.0579)& ($\uparrow$ 1.86) &	($\uparrow$ 0.0227) &	($\uparrow$ 9.38)	 & ($\uparrow$ 0.0720)	 & ($\uparrow$ 0.89) &	($\uparrow$ 0.0288) &	($\uparrow$ 5.61)	 & ($\uparrow$ 0.0453)   \\\hline
\end{tabular}
\end{table*}

To evaluate the transferability of EEGM2's learned representations, we conducted cross-domain experiments. Since the DriverDistraction dataset contains the largest number of samples, we adopted a transfer learning strategy in which EEGM2 is first pre-trained on DriverDistraction and then fine-tuned on the remaining Emotiv datasets. For comparison, we also include results from in-domain pre-training, where EEGM2 is trained directly on each target dataset.

As shown in Table~\ref{tab:6}, EEGM2 significantly outperforms random initialization in both in-domain and cross-domain settings. When pre-trained on DriverDistraction, EEGM2 retains transferable knowledge that leads to consistent performance gains across diverse downstream tasks, including eye state classification (Crowdsourced), mental workload estimation (STEW), alpha wave detection (Alpha), and attention classification. Notably, cross-domain pre-training yields an average improvement of 2.75\% in accuracy and 0.027 in AUROC, demonstrating EEGM2's strong generalization ability across EEG-based tasks. In-domain pre-training still achieves the best overall performance, highlighting the importance of distributional alignment between the pre-training and downstream datasets. Specifically, in-domain pre-training results in a 5.61\% accuracy gain and a 0.0453 AUROC improvement over training from scratch. These results confirm that EEGM2 functions as an effective EEG representation learner, capable of transferring knowledge across varied cognitive and neurological tasks. 

\subsection{Memory Usage \& Inference Speed} \label{sec:4.3}
\begin{table}
\centering
\caption{State-of-the-art EEG models and their size.}
\label{tab:7}
\begin{tabular}{l l c}
\hline
\textbf{Reference}    & \textbf{Model} & \textbf{Model Size} \\ \hline
\cite{yang2024biot}   & BIOT   &  3.2M          \\ 
\cite{chien2211maeeg} & MAEEG  &  2.5M          \\ 
\cite{kostas2021bendr}& BENDR  &  33M           \\ 
\cite{eeg2rep2024}    & EEG2Rep&  0.1M           \\ 
This paper & EEGM2 (Light) & 0.25M \\ 
This paper & EEGM2-S5   & 4.5M  \\ 
This paper & EEGM2   & 4.5M  \\ \hline
\end{tabular}
\end{table}

In this section, we analyze the memory usage and inference speed of recent Transformer-based self-supervised EEG models, alongside the proposed EEGM2, through a simulation experiment.

\noindent\textbf{Related Work:}  
Table~\ref{tab:7} summarizes the number of trainable parameters for each model under the column ``Model Size.'' For comparison, we include EEGM2-S5, a variant of EEGM2 in which the Mamba-2 blocks are replaced with standard Transformer blocks (see Section~\ref{sec:variant}). We also include EEGM2 (Light), a lightweight version of EEGM2, designed for efficient deployment under constrained computational resources. BIOT~\cite{yang2024biot} employs a linear Transformer architecture, enabling efficient modeling of complex token interactions while maintaining linear time complexity. MAEEG~\cite{chien2211maeeg} and BENDR\cite{kostas2021bendr} share a Transformer-based backbone with masking mechanisms but differ in their learning objectives: BENDR combines contrastive and predictive losses to enrich EEG feature learning, while MAEEG adopts a masked autoencoder (MAE) approach, reconstructing masked portions of the EEG signal. EEG2Rep~\cite{eeg2rep2024} uses predictive self-supervised learning with temporal masking to enhance representation quality. This comparison provides insight into the computational efficiency and scalability of EEGM2 relative to other state-of-the-art Transformer-based approaches.

\noindent\textbf{Simulation Setup:}  
The simulation experiment evaluates the performance of various EEG SSL models, including the proposed EEGM2, on a simulated 16-channel EEG input with sequence lengths ranging from 50 to 12,000 time steps. For fair comparison, models that perform downsampling before input are modified to accept full-resolution sequences. Each model undergoes a warm-up phase consisting of 15 inference runs to stabilize GPU performance prior to measurement. All experiments are conducted on a single NVIDIA RTX 6000 Ada GPU with a maximum memory capacity of 51,546~MB. For each sequence length, two metrics are recorded: \textbf{(1)} \textit{Memory usage} (in MB), defined as the peak GPU memory consumption during model inference, and \textbf{(2)} \textit{Inference speed} (in samples/ms), computed as the reciprocal of the average per-sample inference time. The inference time is averaged over 10 independent runs to reduce variance. We compare Transformer-based models, Mamba-based models, and hybrid architectures under identical simulation conditions to assess their scalability and efficiency. The results are visualized as memory usage and inference speed curves plotted against input sequence length.

\begin{figure}
  \centering
  \begin{subfigure}[b]{0.49\linewidth} 
    \centering
    \includegraphics[width=\linewidth]{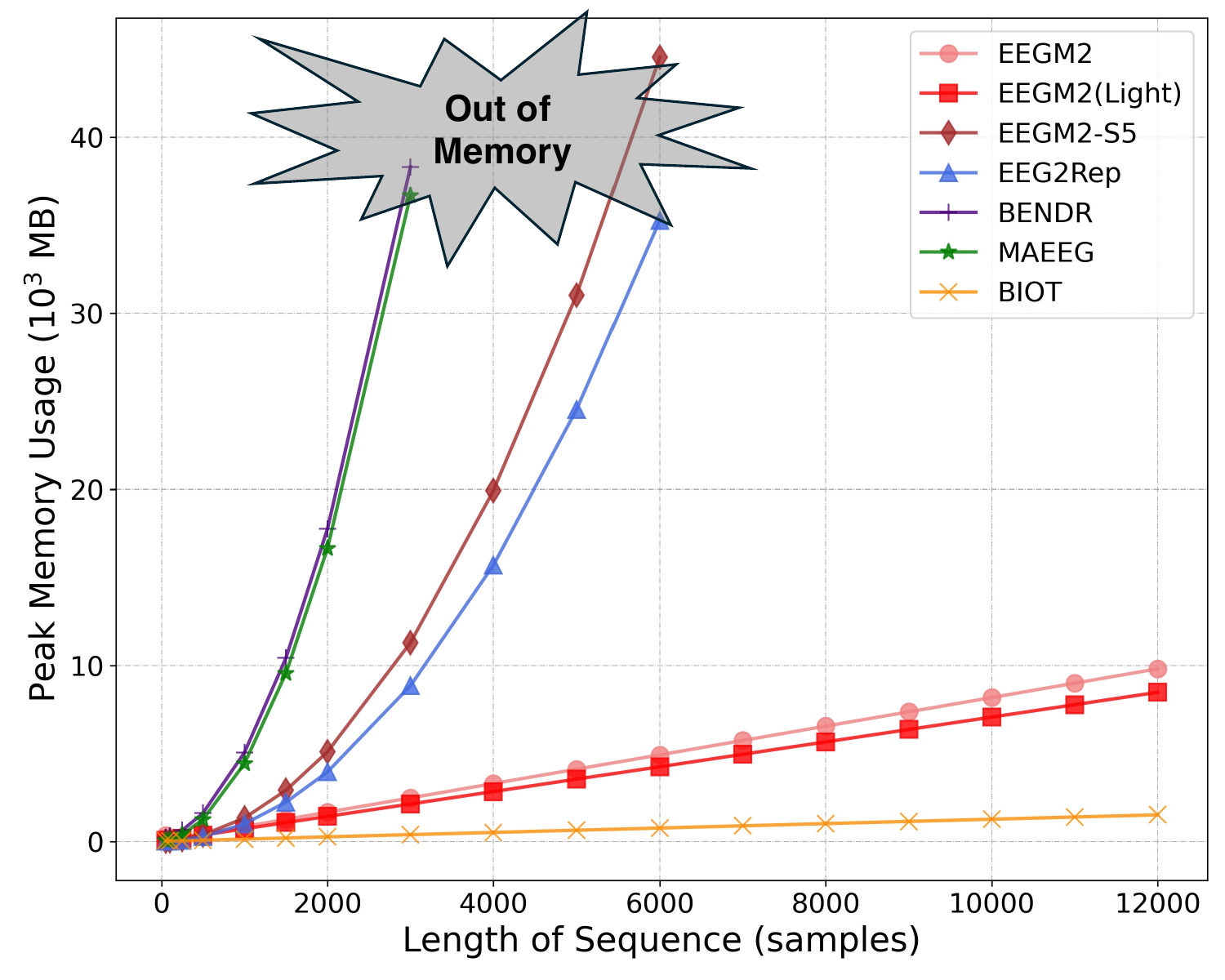}
    \caption{Memory Usage.}
    \label{fig:3a}
  \end{subfigure}
  \hfill 
  \begin{subfigure}[b]{0.49\linewidth}
    \centering
    \includegraphics[width=\linewidth]{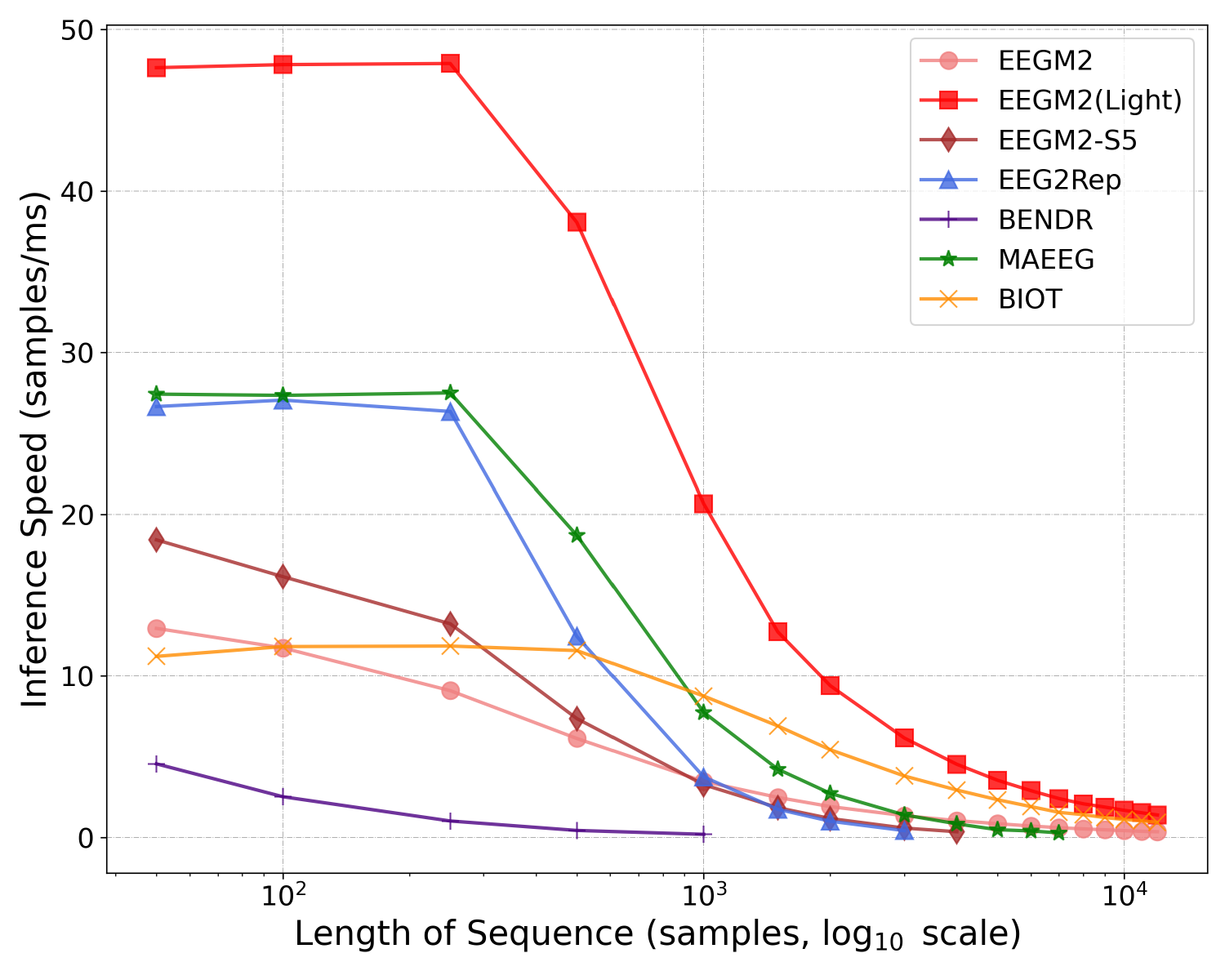}
    \caption{Inference Speed.}
    \label{fig:3b}
  \end{subfigure}
    \caption{Memory usage and inference speed across varying sequence lengths. }
  \label{fig:3}
\end{figure}

\noindent\textbf{Memory Usage:} As shown in Fig.~\ref{fig:3}(a), parameter count is not the dominant factor influencing memory usage. For instance, EEG2Rep has only 0.1M parameters but runs out of memory when the sequence length exceeds 6000. In contrast, EEGM2, powered by the Mamba-2 block, exhibits a linear growth in memory usage with increasing sequence length. Although EEGM2 (Light) reduces parameter size to 0.25M, its memory usage decreases only marginally, indicating that memory consumption is primarily determined by computational complexity rather than model size. MAEEG and BENDR, both based on Transformer architectures, show similar memory profiles, with memory usage scaling quadratically due to the \(\mathcal{O}(N^2)\) complexity of standard self-attention. As a result, they exceed memory limits beyond sequence lengths of approximately 3000. BIOT achieves the lowest memory usage owing to its linear Transformer mechanism and simplified block design. Instead of standard attention, BIOT uses a linear Transformer layer combined with a lightweight feedforward network and residual connections, thereby avoiding the quadratic cost and significantly reducing memory overhead. However, BIOT's most time-consuming component is its Fourier transformer-based preprocessing, which impacts memory usage but results in slower inference speed, as shown in Fig.~\ref{fig:3}(b). EEGM2-S5, a variant of EEGM2 in which the Mamba-2 block is replaced by a standard Transformer block, also fails beyond 6000 time steps, further highlighting the memory efficiency advantage of the Mamba-2 block.

\noindent\textbf{Inference Speed:} As shown in Fig.~\ref{fig:3}(b), we evaluate model efficiency in terms of inference speed measured by samples per millisecond (samples/ms), which reflects their suitability for deployment on resource-constrained edge BCI devices. A base-10 logarithmic scale is used for plotting inference speed to improve visual clarity across different sequence lengths, allowing for consistent comparison of model performance trends. Unlike memory usage, inference speed is more sensitive to parameter size and preprocessing overhead. Transformer-based models, including MAEEG, EEG2Rep, and EEGM2-S5, suffer from rapidly declining inference speeds as sequence length increases due to their quadratic complexity. BENDR, with its large parameter size (33M), exhibits the slowest inference among all models. Although BIOT employs a linear Transformer mechanism, its inference speed remains limited, largely due to its Fourier-based preprocessing and a relatively large parameter count (3.2M). In contrast, EEGM2 (Light) achieves the highest inference speed across all sequence lengths. This efficiency stems from its lightweight architecture (0.25M parameters) and use of the Mamba-2 block, which enables faster processing without sacrificing performance. These results demonstrate EEGM2 (Light)'s potential as a practical and scalable solution for real-time BCI applications, offering both speed and accuracy for downstream EEG tasks.

\section{Conclusion}
We propose EEGM2, a Mamba-2-based self-supervised framework for efficient modeling of EEG sequences across both short and long durations. By integrating structured state-space modeling, a temporal-spectral loss, and multi-scale receptive field embeddings, EEGM2 captures long-range dependencies with high computational efficiency. Experiments on multiple EEG datasets show that EEGM2 outperforms Transformer-based baselines, demonstrating strong generalization and transferability in downstream tasks. Simulation results further confirm its advantages in memory usage and inference speed. Additionally, EEGM2 (Light), a lightweight variant with 18$\times$ fewer parameters, maintains competitive performance, offering a practical alternative for resource-constrained scenarios. Overall, EEGM2 provides a scalable and efficient solution for EEG representation learning.

\bibliographystyle{IEEEtran}
\bibliography{IEEEfull,IEEEref}

\appendix
\subsection{Emotiv Dataset} \label{app:1}
\noindent\textbf{Attention} dataset  was collected through an experiment where subjects completed four tasks—two visual and two auditory—designed to assess attention in classifying repeated stimuli. In visual tasks, participants viewed four-digit numbers and clicked when the same number appeared consecutively, with a total duration of 640 seconds. In auditory tasks, they listened to three words and clicked when a word was repeated in sequence, lasting 540 seconds. Each subject completed a total stimulus time of 19 minutes and 40 seconds. Data were recorded using a 14-channel Emotiv Epoc headset, generating multivariate time-series data. After preprocessing and manual labeling, data from 31 subjects were collected, with 4 excluded due to poor quality.

\noindent\textbf{Crowdsourced} is a publicly available dataset \cite{williams2023crowdsourced} collected while participants performed a resting-state task, alternating between two-minute intervals with eyes open and eyes closed. Among the 60 participants, only 13 successfully completed both conditions using 14-channel EPOC$+$, EPOC X, and EPOC devices. The data was originally recorded at 2048 Hz and later downsampled to 128 Hz. The raw EEG recordings from these 13 participants, along with pre-processing, analysis, and visualization scripts, are publicly accessible on the Open Science Framework (OSF).

\noindent\textbf{DriverDistraction} dataset was obtained by recording EEG brain activity from 17 participants while they engaged in a driving simulation for around 40 minutes. During the simulation, participants carried out various distraction tasks, which can be categorized into three main types: (1) conversing with a passenger, (2) interacting with a mobile phone (including texting and calling), and (3) engaging in problem-solving activities. EEG signals were captured at a sampling rate of 128 Hz using the Emotiv Epoc EEG headset, which records data from 14 channels. The resulting dataset is a multivariate time series with 14 input variables and approximately 5.5 million records. Each time point in the dataset was manually labeled according to the specific activity being performed.

\noindent\textbf{STEW} dataset is a publicly available dataset \cite{lim2018stew} that consists of raw EEG recordings collected from 48 participants who took part in a multitasking workload experiment using the SIMKAP multitasking test. Prior to the test, baseline brain activity at rest was also recorded. EEG signals were captured using a 14-channel Emotiv EPOC headset at a sampling rate of 128 Hz, resulting in 2.5 minutes of recorded data per participant. After each stage of the experiment, participants assessed their perceived mental workload on a scale from 1 to 9, with these ratings stored in a separate file. Additionally, the dataset includes binary class labels, where workload ratings greater than 4 are categorized as high, while ratings of 4 or below are classified as low. These labels are utilized for specific analytical purposes. The STEW dataset is available upon request via IEEE DataPort.

\subsection{Ablation Study - Long-Sequence Modeling} \label{app:2}
\begin{figure}
  \centering
  \begin{subfigure}[b]{0.48\linewidth} 
    \centering
    \includegraphics[width=\linewidth]{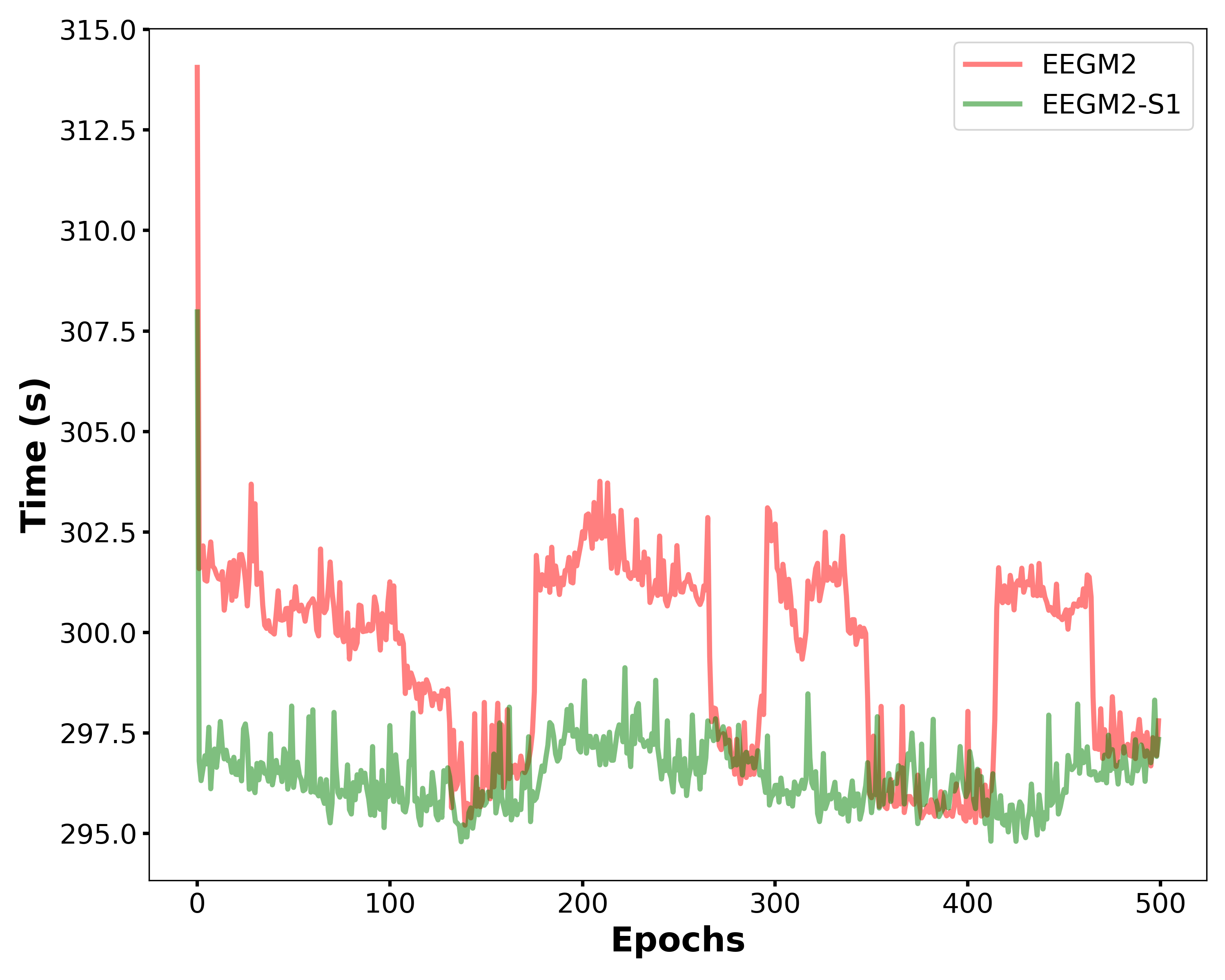}
    \caption{EEGM2 and EEGM2-S1.}
    \label{fig:app1}
  \end{subfigure}
  \hfill 
  \begin{subfigure}[b]{0.48\linewidth}
    \centering
    \includegraphics[width=\linewidth]{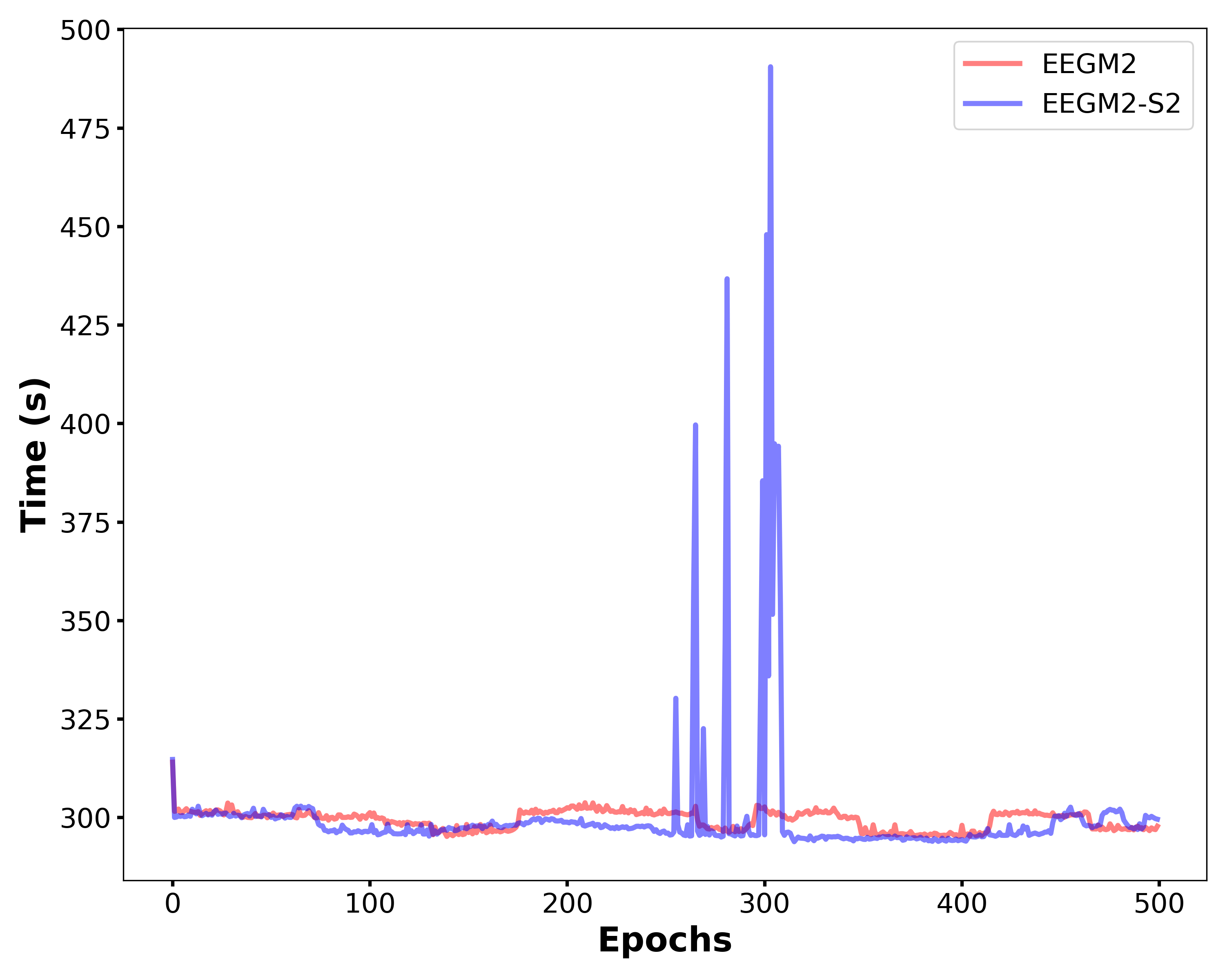}
    \caption{EEGM2 and EEGM2-S2.}
    \label{fig:app2}
  \end{subfigure}
  \hfill 
  \begin{subfigure}[b]{0.48\linewidth}
    \centering
    \includegraphics[width=\linewidth]{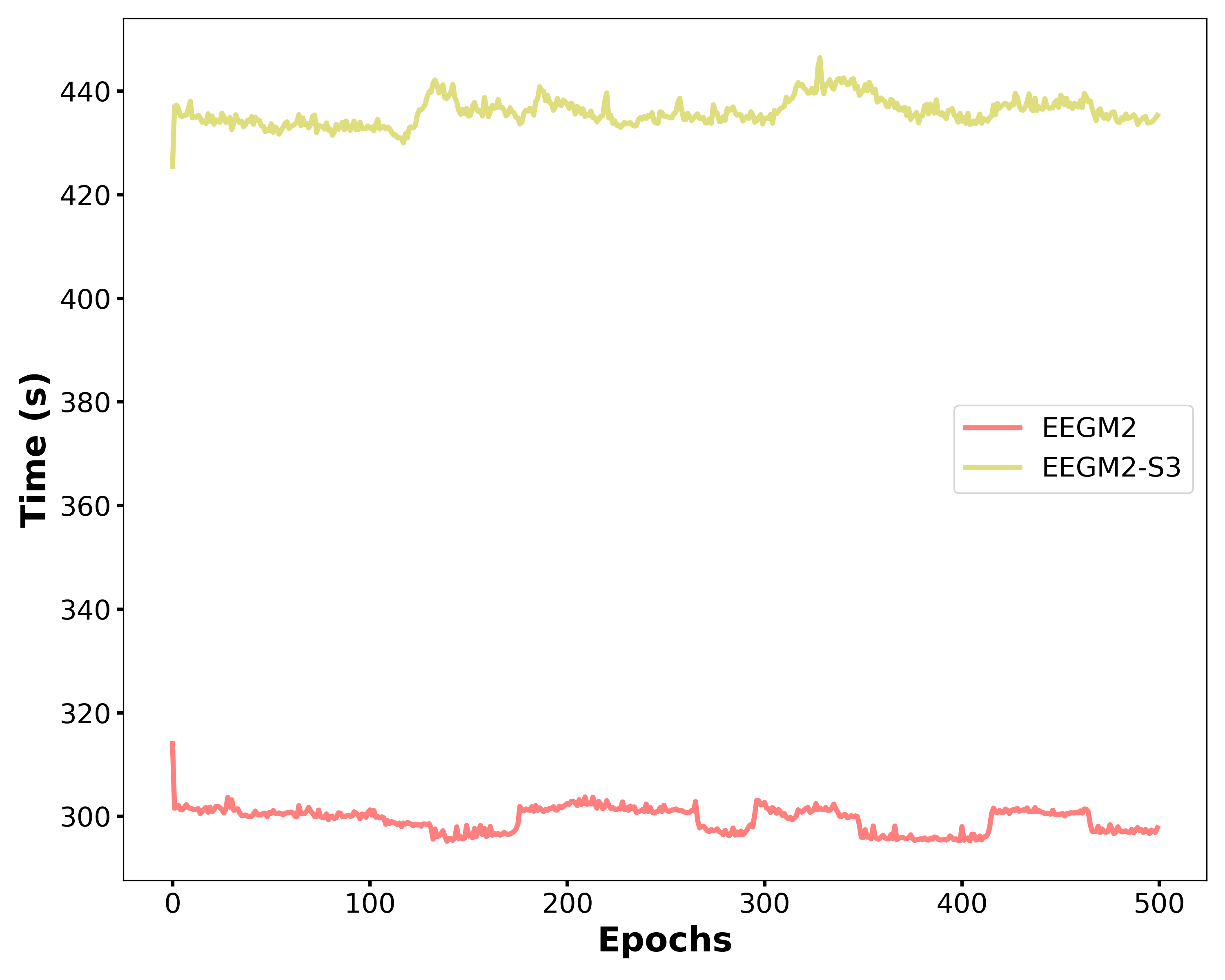}
    \caption{EEGM2 and EEGM2-S3.}
    \label{fig:app3}
  \end{subfigure}
  \hfill 
  \begin{subfigure}[b]{0.48\linewidth}
    \centering
    \includegraphics[width=\linewidth]{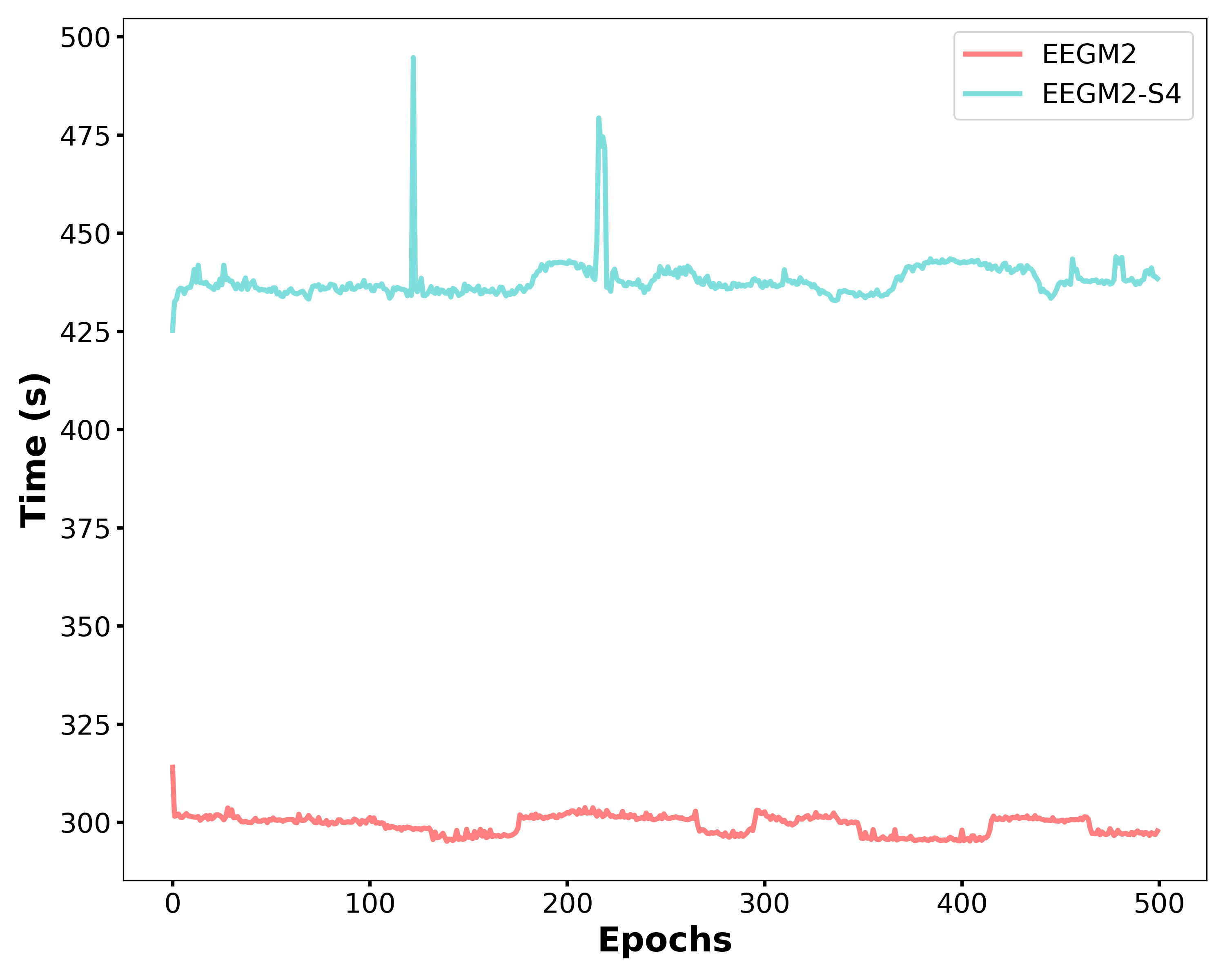}
    \caption{EEGM2 and EEGM2-S4.}
    \label{fig:app4}
  \end{subfigure}
  \caption{Training time of EEGM2 and its variants across four different settings.}
  \label{fig:4}
\end{figure}

Fig.\ref{fig:4} provides additional insights into the training time across 500 epochs for EEGM2 and its ablation variants. In Fig.\ref{fig:4}(a), we observe that the inclusion of the multi-scale receptive field (EEGM2 vs. EEGM2-S1) incurs only minimal computational overhead, as evidenced by the small difference in training time. Similarly, Fig.~\ref{fig:4}(b) shows that EEGM2 and EEGM2-S2 (which replaces the temporal-spectral loss with L1 loss) exhibit comparable training durations. The occasional spikes in EEGM2-S2 may result from transient GPU scheduling variance or momentary hardware resource contention, rather than differences in the loss function itself.

In contrast, Figs.~\ref{fig:4}(c) and \ref{fig:4}(d) reveal that replacing Mamba-2 with Mamba-1 (EEGM2-S3 and EEGM2-S4) leads to a notable increase in training time, indicating a higher computational cost. These findings highlight the efficiency advantage of the Mamba-2 block, which enables faster training while preserving model capacity.

\end{document}